\documentclass[sigconf]{acmart}

\usepackage{graphicx}
\usepackage{subcaption}
\usepackage{amsmath}
\usepackage{amsfonts}
\usepackage[ruled,vlined]{algorithm2e}
\usepackage{bm}
\usepackage{multirow}
\usepackage{bbm}

\usepackage{natbib}
\usepackage{fancyhdr}

\DeclareMathOperator*{\argmin}{arg\,min}

\AtBeginDocument{%
  \providecommand\BibTeX{{%
    \normalfont B\kern-0.5em{\scshape i\kern-0.25em b}\kern-0.8em\TeX}}}

\copyrightyear{2021}
\acmYear{2021}
\setcopyright{acmcopyright}\acmConference[KDD '21]{Proceedings of the 27th ACM
SIGKDD Conference on Knowledge Discovery and Data Mining}{August 14--18,
2021}{Virtual Event, Singapore}
\acmBooktitle{Proceedings of the 27th ACM SIGKDD Conference on Knowledge
Discovery and Data Mining (KDD '21), August 14--18, 2021, Virtual Event, Singapore}
\acmPrice{15.00}
\acmDOI{10.1145/3447548.3467274}
\acmISBN{978-1-4503-8332-5/21/08}

\begin{document}
\fancyhead{}

\title{S-LIME: Stabilized-LIME for Model Explanation}

\author{Zhengze Zhou}
\affiliation{%
  \institution{Cornell University}
  \city{Ithaca}
  \state{New York}
  \country{USA}
  \postcode{14850}}
\email{zz433@cornell.edu}

\author{Giles Hooker}
\affiliation{%
  \institution{Cornell University}
  \city{Ithaca}
  \state{New York}
  \country{USA}
  \postcode{14850}}
\email{gjh27@cornell.edu}

\author{Fei Wang}
\affiliation{%
  \institution{Weill Cornell Medicine}
  \city{New York City}
  \state{New York}
  \country{USA}
  \postcode{10065}}
\email{few2001@med.cornell.edu}

\begin{abstract}
An increasing number of machine learning models have been deployed in domains with high stakes such as finance and healthcare. Despite their superior performances, many models are black boxes in nature which are hard to explain. There are growing efforts for researchers to develop methods to interpret these black-box models. Post hoc explanations based on perturbations, such as LIME \citep{ribeiro2016should}, are widely used approaches to interpret a machine learning model after it has been built. This class of methods has been shown to exhibit large instability, posing serious challenges to the effectiveness of the method itself and harming user trust. In this paper, we propose S-LIME, which utilizes a hypothesis testing framework based on central limit theorem for determining the number of perturbation points needed to guarantee stability of the resulting explanation. Experiments on both simulated and real world data sets are provided to demonstrate the effectiveness of our method. 
\end{abstract}




\maketitle

\section{Introduction}

Data Mining and machine learning models have been widely deployed for decision making in many fields, including criminal justice \citep{zeng2015interpretable} and healthcare \citep{rajkomar2018scalable, miotto2018deep}. However, many models act as ``black boxes" in that they only provide predictions but with little guidance for humans to understand the process. It has been a desiderata to develop approaches for understanding these complex models, which can help increase user trust \citep{ribeiro2016should}, assess fairness and privacy \citep{angwin2016machine, doshi2017towards}, debug models \citep{koh2017understanding} and even for regulation purposes \citep{goodman2017european}.

Model explanation methods can be roughly divided into two categories \cite{du2019techniques,wang2020should}: intrinsic explanations and post hoc explanations. Models with intrinsically explainable structures include linear models, decision trees \citep{breiman1984classification}, generalized additive models \citep{hastie1990generalized}, to name a few. Due to complexity constraints, these models are usually not powerful enough for modern tasks involving heterogeneous features and enormous numbers of samples. 

Post hoc explanations, on the other hand, provide insights after a model is trained. These explanations can be either model-specific, which are typically limited to specific model classes, such as split improvement for tree-based methods \citep{zhou2019unbiased} and saliency maps for convolutional networks \citep{simonyan2013deep}; or model-agnostic that do not require any knowledge of the internal structure of the model being examined, where the analysis is often conducted by evaluating model predictions on a set of perturbed input data. LIME \citep{ribeiro2016should} and SHAP \citep{lundberg2017unified} are two of the most popular model-agnostic explanation methods. 

Researchers have been aware of some drawbacks for post hoc model explanation. \cite{hooker2019please} showed that widely used permutation importance can produce diagnostics that are highly misleading due to extrapolation. \cite{ghorbani2019interpretation} demonstrated how to generate adversarial perturbations that produce perceptively indistinguishable inputs with the
same predicted label, yet have very different interpretations. \cite{aivodji2019fairwashing} showed that explanation algorithms can be exploited to systematically rationalize decisions taken by an unfair black-box model. \cite{rudin2019stop} argued against using post hoc explanations as these methods can provide explanations that are not faithful to what the original model computes.

In this paper, we focus on post hoc explanations based on perturbations \citep{ribeiro2016should}: one of the most popular paradigm for designing model explanation methods. We argue that the most important property of any explanation technique is \emph{stability} or \emph{reproducibility}: repeated runs of the explanation algorithm under the same conditions should ideally yield the same results. Unstable explanations provide little insight to users as how the model actually works and are considered unreliable. Unfortunately, LIME is not always stable. \cite{zhang2019should} separated and investigated sources of instability in LIME. \cite{visani2020optilime} highlighted a
trade-off between explanation’s stability and adherence and propose a framework to maximise stability. \cite{lee2019developing} improved the sensitivity of LIME by averaging multiple output weights for individual images. 

We propose a hypothesis testing framework based on a central limit theorem for determining the number of perturbation samples required to guarantee stability of the resulting explanation. Briefly, LIME works by generating perturbations of a given instance and learning a sparse linear explanation, where the sparsity is usually achieved by selecting top features via LASSO \citep{tibshirani1996regression}. LASSO is known to exhibit early occurrence of false discoveries \citep{meinshausen2010stability, su2017false} which, combined with the randomness introduced in the sampling procedure, results in practically-significant levels of instability. We carefully analyze the Least Angle Regression (LARS) \citep{efron2004least} for generating the LASSO path and quantify the aymptotics for the statistics involved in selecting the next variable. Based on a hypothesis testing procedure, we design a new algorithm call S-LIME (Stabilized-LIME) which can automatically and adaptively determine the number of perturbations needed to guarantee a stable explanation. 

In the following, we review relevant background on LIME and LASSO along with their instability in Section \ref{backgound}. Section \ref{asymptotics} statistically analyzes the asymptotic distribution of the statistics which is at the heart of variable selection in LASSO. Our algorithm S-LIME is introduced in Section \ref{algo}. Section \ref{empirical} presents empirical studies on both simulated and real world data sets. We conclude in Section \ref{conclusion} with some discussions. 

\section{Background}\label{backgound}

In this section, we review the general framework for constructing {\em post hoc} explanations based on perturbations using Local Interpretable Model-agnostic Explanations (LIME) \citep{ribeiro2016should}. We then briefly discuss LARS and LASSO, which are the internal solvers for LIME to achieve the purpose of feature selection. We illustrate LIME's instability with toy experiments. 

\subsection{LIME}

Given a black box model $f$ and a target point $\bm{x}$ of interest, we would like to understand the behavior of the model locally around $\bm{x}$. No knowledge of $f$'s internal structure is available but we are able to query $f$ many times. LIME first samples around the neighborhood of $\bm{x}$, query the black box model $f$ to get its predictions and form a pseudo data sets $\mathcal{D} = \{(\bm{x}_1, y_1), (\bm{x}_2, y_2), \ldots, (\bm{x}_n, y_n)\}$ with $y_i = f(\bm{x}_i)$ and a hyperparameter $n$ specifying the number of perturbations. The model $f$ can be quite general as regression ($y_i \in \mathcal{R}$) or classification ($y_i \in \{0, 1\}$ or $y_i \in [0, 1]$ if $f$ returns a probability). A model $g$ from some interpretable function spaces $G$ is chosen by solving the following optimization
\begin{equation}\label{kLASSO}
    \argmin_{g \in G}L(f, g, \pi_{\bm{x}})  + \Omega(g)
\end{equation}
where
\begin{itemize}
    \item $\pi_{\bm{x}}(\bm{z})$ is a proximity measure between a perturbed instance $\bm{z}$ to $\bm{x}$, which is usually chosen to be a Gaussian kernel. 
    \item $\Omega(g)$ measures complexity of the explanation $g \in G$. For example, for decision trees $\Omega(g)$ can be the depth of the tree, while for linear models we can use the number of non-zero weights. 
    \item $L(f, g, \pi_{\bm{x}})$ is a measure of how unfaithful $g$ is in approximating $f$ in the locality defined by $\pi_{\bm{x}}$.
\end{itemize}
\cite{ribeiro2016should} suggests a procedure called k-LASSO for selecting top $k$ features using LASSO. In this case, $G$ is the class of linear models with $g = \bm{\omega}_g \cdot \bm{x}$, $L(f, g, \pi_{\bm{x}}) = \sum_{i = 1}^n\pi_{\bm{x}}(\bm{x}_i)(y_i - g(\bm{x}_i))^2$ and $\Omega = \infty \mathbbm{1}[||\omega_g||_0 > k]$. Under this setting, (\ref{kLASSO}) can be approximately solved by first selecting K features with LASSO (using the regularization
path) and then learning the weights via least square \citep{ribeiro2016should}.

\textcolor{black}{We point out here the resemblance between post hoc explanations and knowledge distillation \citep{bucilua2006model,hinton2015distilling}; both involve obtaining predictions from the original model, usually on synthetic examples, and using these to train a new model. Differences lie in both the scope and intention in the procedure. Whereas LIME produces interpretable models that apply closely to the point of interest, model distillation is generally used to provide a global compression of the model representation in order to improve both computational and predictive performance \citep{gibbons2013computerized,menon2020distillation}. Nonetheless, we might expect that distillation methods to also exhibit the instability described here; see \citep{zhou2018approximation} which documents instability of decision trees used to provide global interpretation.}


\subsection{LASSO and LARS}

Even models that are ``interpretable by design" can be difficult to understand, such as a deep decision tree containing hundreds of leaves, or a linear model that employs many features with non-zero weights. For this reason LASSO \citep{tibshirani1996regression}, which automatically produces sparse models, is often the default solver for LIME. 

Formally, suppose $\mathcal{D} = \{(\bm{x}_1, y_1), (\bm{x}_2, y_2), \ldots, (\bm{x}_n, y_n)\}$ with $\bm{x}_i = (x_{i1}, x_{i2}, \ldots, x_{ip})$ for $1 \leq i \leq n$, LASSO solves the following optimization problem:
\begin{equation}\label{LASSO}
    \hat{\beta}^{LASSO} = \argmin_{\beta}\left\{\sum_{i=1}^n(y_i - \beta_0 - \sum_{j = 1}^px_{ij}\beta_j)^2 + \lambda\sum_{j=1}^p|\beta_j|\right\}
\end{equation}
where $\lambda$ is the multiplier for $l_1$ penalty. (\ref{LASSO}) can be efficiently solved via a slight modification of the LARS algorithm \citep{efron2004least}, which gives the entire LASSO path as $\lambda$ varies. This procedure is described in Algorithm \ref{a_lars} and \ref{a_LASSO} below \citep{friedman2001elements}, where we denote $\bm{y} = (y_1, y_2, \ldots, y_n)$ and assume $n > p$.

\begin{algorithm}[]
\SetAlgoLined
\begin{enumerate}
    \item Standardize the predictors to have zero mean and unit norm. Start with residual $\bm{r} = \bm{y} - \bm{\bar{y}}, \beta_1, \beta_2, \ldots, \beta_p = 0$.
    \item Find the predictor $\bm{x}_{\cdot j}$ most correlated with $\bm{r}$, and move $\beta_j$ from 0 towards its least-squares coefficient $\langle\bm{x}_{\cdot j}, \bm{r}\rangle$, until some other competitors $\bm{x}_{\cdot k}$ has as much correlation with the current residual as does $\bm{x}_{\cdot j}$.
    \item Move $\beta_j$ and $\beta_k$ in the direction defined by their joint least squares coefficient of the current residual on $(\bm{x}_{\cdot j}, \bm{x}_{\cdot k})$, until some other competitors $\bm{x}_{\cdot l}$ has as much correlation with the current residual. 
    \item Repeat step 2 and 3 until all $p$ predictors have been entered, at which point we arrive at the full least squares solution. 
\end{enumerate}
 \caption{Least Angle Regression (LARS)}
 \label{a_lars}
\end{algorithm}

\begin{algorithm}[]
\SetAlgoLined
\begin{itemize}
  \item[3a.]  In step 3 of Algorithm \ref{a_lars}, if a non-zero coefficient hits zero, drop the corresponding variable from the active set of variables and recompute the current joint least squares direction.
\end{itemize}
 \caption{LASSO: Modification of LARS}
 \label{a_LASSO}
\end{algorithm}

Both Algorithm \ref{a_lars} and \ref{a_LASSO} can be easily modified to incorporate a weight vector $\bm{\omega} = (\omega_1, \omega_2, \ldots, \omega_n)$ on the data set $\mathcal{D}$, by transforming it to $\mathcal{D} = \{(\sqrt{\omega_1}\bm{x}_1, \sqrt{\omega_1}y_1), (\sqrt{\omega_2}\bm{x}_2, \sqrt{\omega_2}y_2), \ldots, (\sqrt{\omega_n}\bm{x}_n, \sqrt{\omega_n}y_n)\}$.

\subsection{Instability with LIME}\label{lime_insta}

\begin{figure}[h]
    \centering
    \includegraphics[width = \linewidth]{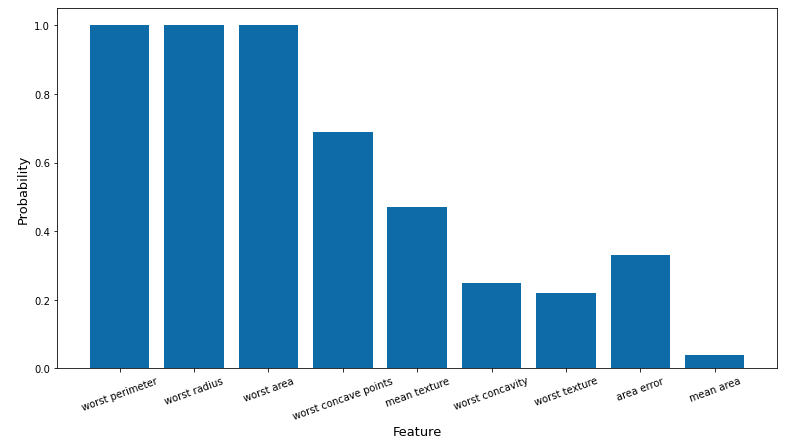}
    \caption{Empirical selection probability for features in Breast Cancer Data. The black box model is a random forests classifier with 500 trees. LIME is run 100 times on a randomly selected test point and top 5 features are selected via LASSO. }
    \label{fig:breast_feature}
\end{figure}

Both \cite{zhang2019should} and \cite{zafar2019dlime} have demonstrated that the random generation of perturbations results in instability in the generated explanations.  We apply LIME on Breast Cancer Data (see Section 
\ref{breast} for details) to illustrate of this phenomenon. A random forests \citep{breiman2001random} with 500 trees is built as the black box model, and we apply LIME to explain the prediction of a randomly selected test point multiple times. Each time 1000 synthetic data are generated around the test point and top 5 features are selected via LASSO. We repeat the experiment 100 times and calculate the empirical selection probability of features. The result is shown in Figure \ref{fig:breast_feature}. 

We can see that across 100 repetitions, only three features are consistently selected by LIME while there is considerable variability in the remaining features. Note that this does not consider the order of the features entered: even the top three features exhibit different orderings in the selection process.

This experiment illustrates an important weakness of LIME: its instability or irreproducibility. If repeated runs using the \emph{same} explanation algorithm on the \emph{same} model to interpret the \emph{same} data point yield different results, the utility of the explanation is brought into question. The instability comes from the randomness introduced when generating synthetic samples around the input, and the $l_1$ penalty employed in LASSO further increases the chance of selecting spurious features \citep{su2018first}. In Appendix \ref{app:LASSO} we show the instability with LASSO using a simple linear model. 

One way to stabilize the LIME model is to use a larger corpus of the synthetic data, but it is difficult to determine how much larger \emph{as a priori} without repeated experiments. In the next section, we examine how feature selection works in LASSO and LARS, and then design a statistically justified approach to automatically and adaptively determine the number of perturbations required to guarantee stability.

\section{Asymptotic Properties of LARS Decisions}\label{asymptotics}

Consider at any given step when LARS needs to choose a new variable to enter the model. With sample size of $n$, let the current residuals be given by $\bm{r} = (r_1, r_2, \ldots, r_n)$, and two candidate variables being $\bm{x_{\cdot i}} = (x_{1i}, x_{2i}, \ldots, x_{ni})$ and $\bm{x_{\cdot j}} = (x_{1j}, x_{2j}, \ldots, x_{nj})$ \textcolor{black}{ where we assume the predictors have been standardized to have zero mean and unit norm.  LARS chooses the predictor that has the highest (absolute) correlation with the residuals to enter the model. Equivalently, one needs to compare $
\hat{c}_1 = \frac{1}{n}\sum_{t = 1}^n r_tx_{ti}$ with $\hat{c}_2 = \frac{1}{n}\sum_{t = 1}^n r_tx_{tj}$. We use $\hat{c}_1$ and $\hat{c}_2$ to emphasize these are finite sample estimates, and our purpose is to obtain the probability that their order would be different if the query points were regenerated. To that end, we introduce uppercase symbols $\bm{R}, \bm{X_{\cdot i}}, \bm{X_{\cdot j}}$ to denote the corresponding random variables of the residuals and two covariates; these are distributed according to the current value of the coefficients in the LASSO path and we seek to generate enough data to return the same ordering as the expected values $c_1 = E(\bm{R} \cdot \bm{X_{\cdot i}})$ and $c_2 = E(\bm{R} \cdot \bm{X_{\cdot j}})$ with high probability. Our algorithm is based on pairwise comparisons between candidate features; we therefore consider the decision between two covariates in this section, and extensions to more general cases involving multiple pairwise comparisons will be discussed in Section \ref{algo}. } 


By the multivariate Central Limit Theorem (CLT), we have 
\begin{align*}
    \sqrt{n} \left(\begin{bmatrix}
    \hat{c}_1 \\
    \hat{c}_2
    \end{bmatrix} -
    \begin{bmatrix}
    c_1 \\
    c_2
    \end{bmatrix} \right) \longrightarrow
    N(0, \, \Sigma),
\end{align*}

where 
\begin{align*}
    \Sigma  = \mbox{cov} \begin{bmatrix}
    \bm{R} \cdot \bm{X_{\cdot i}}\\
    \bm{R} \cdot \bm{X_{\cdot j}}
    \end{bmatrix} = \begin{bmatrix}
    \sigma_{11}^2 & \sigma_{12}^2 \\
    \sigma_{21}^2 & \sigma_{22}^2
    \end{bmatrix}.
\end{align*}

Without loss of generality we assume $\hat{c}_1 > \hat{c}_2 > 0$. In general if the correlation is negative, we can simply negate the corresponding feature values for all the calculations involved in this section. Let $\hat{\Delta}_n = \hat{c}_1 - \hat{c}_2$ and $\Delta_n = c_1 - c_2$. Consider function $f(a_1, a_2) = a_1 - a_2$. Delta method implies that 
\begin{align*}
    \sqrt{n} \left(f\left(\begin{bmatrix}
    \hat{c}_1 \\
    \hat{c}_2
    \end{bmatrix}\right) - \left(f
    \begin{bmatrix}
    c_1 \\
    c_2
    \end{bmatrix}\right) \right) \longrightarrow
    N(0, \, \sigma_{11}^2 + \sigma_{22}^2 - \sigma_{12}^2 - \sigma_{21}^2).
\end{align*}
Or approximately,
\begin{equation}\label{eqn:diff}
    \hat{\Delta}_n - \Delta_n \sim N\Big(0, \, \frac{ \hat{\sigma}_{11}^2 + \hat{\sigma}_{22}^2 - \hat{\sigma}_{12}^2 - \hat{\sigma}_{21}^2}{n}\Big)
\end{equation}
\textcolor{black}{where the variance estimates are estimated from the empirical covariance of the values $r_t x_{t i}$ and $r_t x_{t j}$, $t = 1,\ldots,n$.}


In similar spirits of \cite{zhou2018approximation}, we assess the probability that $\hat{\Delta}_n > 0$ will still hold in a repeated experiment. Assume we have another independently generated data set denoted by $\{r_t^*, x_{ti}^*, x_{tj}^*\}_{t=1}^{n}$. It follows from (\ref{eqn:diff}) that
$$
    \hat{\Delta}_n^* - \hat{\Delta}_n \sim N\Big(0, \, 2 \cdot \frac{\hat{\sigma}_{11}^2 + \hat{\sigma}_{22}^2 - \hat{\sigma}_{12}^2 - \hat{\sigma}_{21}^2}{n}\Big),
$$
which leads to the approximation that
$$
\hat{\Delta}_n^* \bigg| \left(\hat{\Delta}_n = \hat{c}_1 - \hat{c}_2\right) \sim N\Big(\hat{c}_1 - \hat{c}_2, \, 2 \cdot \frac{\hat{\sigma}_{11}^2 + \hat{\sigma}_{22}^2 - \hat{\sigma}_{12}^2 - \hat{\sigma}_{21}^2}{n}\Big).
$$
In order to control $P(\hat{\Delta}_n^* > 0)$ at a confidence level $1-\alpha$, we need
\begin{eqnarray}
\label{eqn:z}
\hat{c}_1 - \hat{c}_2 > Z_{\alpha} \sqrt{2\frac{\hat{\sigma}_{11}^2 + \hat{\sigma}_{22}^2 - \hat{\sigma}_{12}^2 - \hat{\sigma}_{21}^2}{n}},
\end{eqnarray}
where $Z_{\alpha}$ is the $(1 - \alpha)$-quantile of a standard normal distribution.

For a fixed confidence level $\alpha$ and $n$, suppose we get the corresponding $p$-value $p_n > \alpha$. From (\ref{eqn:z}) we have
$$
\sqrt{n} \frac{\hat{c}_1 - \hat{c}_2}{\sqrt{2(\hat{\sigma}_{11}^2 + \hat{\sigma}_{22}^2 - \hat{\sigma}_{12}^2 - \hat{\sigma}_{21}^2)}} =  Z_{p_n}.
$$
This implied we would need approximately $n'$ samples to get a significant result where
\begin{equation}\label{eqn:newn}
    \sqrt{\frac{n}{n'}} = \frac{Z_{p_n}}{Z_{\alpha}}.
\end{equation}

\section{Stabilized-LIME}\label{algo}

Based on the theoretical analysis developed in Section \ref{asymptotics}, we can run LIME equipped with hypothesis testing at each step when a new variable enters. If the testing result is significant, we continue to the next step; otherwise it indicates that the current sample size of perturbations is not large enough. We thus generate more synthetic data according to Equation (\ref{eqn:newn}) and restart the whole process. \textcolor{black}{Note that we view any intermediate step as \emph{conditioned on} previous obtained estimates of $\hat{\beta}$.} A high level sketch of the algorithm is presented below in Algorithm \ref{a_slime}. 

\begin{algorithm}[]
\SetKwInOut{Input}{Input}\SetKwInOut{Output}{Output}
\SetAlgoLined
\Input{A black box model $f$, data sample to explain $\bm{x}$, initial size for perturbation samples $n_0$, significance level $\alpha$, number of features to select $k$, proximity measure $\pi_{\bm{x}}$.}
\Output{Top $k$ features selected for interpretation. }
Generate $\mathcal{D}$ = $\{n_0$ synthetic samples around $\bm{x}\}$ and calculate weight vector $\bm{\omega}$ using $\pi_{\bm{x}}$\;
Set $n = n_0$\;
 \While{True}{
  Run Algorithm \ref{a_LASSO} on $\mathcal{D}$ with weight $\bm{\omega}$ along with hypothesis testing at each step:
  
  \While{active features less than $k$} {
  \textcolor{black}{Select top two predictors most correlated with the current residual from remaining covariates, with covariance $\hat{c}_1$ and $\hat{c}_2$\;}
  Calculate test statistic:
  $$
  t = \hat{c}_1 - \hat{c}_2 - Z_{\alpha} \sqrt{2\frac{\hat{\sigma}_{11}^2 + \hat{\sigma}_{22}^2 - \hat{\sigma}_{12}^2 - \hat{\sigma}_{21}^2}{n}}\;
  $$
  \eIf{$t >= 0$}{
  Continue with this selection\;
  }{
  Calculate $n' = n * \Big(\frac{Z_{\alpha}}{Z_{p_n}}\Big)^ 2$ and set $n = n'$\;
  Break;
  }  
  }
  \eIf{active features less than $k$}{
    Generate $\mathcal{D}$ = $\{n'$ synthetic samples around $\bm{x}\}$ and calculate weight vector $\bm{\omega}$ using $\pi_{\bm{x}}$\;
  }{
    Return $k$ selected features\;
  }  
 }
 \caption{S-LIME}
 \label{a_slime}
\end{algorithm}

In practice, we may need to set an upper bound on the number of synthetic samples generated (denoted by $n_{max}$), such that whenever the new $n'$ is greater than $n_{max}$, we'll simply set $n = n_{max}$ and go though the outer while loop one last time \emph{without} testing at each step. This can prevent the algorithm from running too long and wasting computation resources in cases where two competing features are equally important in a local neighborhood; for example, if the black box model is indeed locally linear with equal coefficients for two predictors. 

We note several other possible variations of the Algorithm \ref{a_slime}.

\textbf{Multiple testing}. So far we've only considered comparing a pair of competing features (the top two). But when choosing the next predictor to enter the model at step $m$ (with $m - 1$ active features), there are $p - m + 1$ candidate features. We can modify the procedure to select the best feature among all the remaining candidates, by conducting pairwise comparisons between the feature with largest correlation ($\hat{c}_1$) against the rest ($\hat{c}_2, \ldots, \hat{c}_{p - m + 1}$). This is a multiple comparisons problem, and one can use an idea analogous to Bonferroni correction. Mathematically:
\begin{itemize}
    \item Test the hypothesis $H_{i, 0}: \hat{c}_1  \leq \hat{c}_{i}$, $i = 2, \ldots, p - m + 1$. Obtain $p$-values $p_2, \ldots, p_{p - m + 1}$.
    \item Reject the null hypothesis if $\sum_{i = 2}^{p - m + 1}p_i < \alpha$.
\end{itemize}
Although straightforward, this Bonferroni-like correction ignores much of the correlation among these statistics and will result in a conservative estimate. In the experiments, we only conduct hypothesis testing for top two features without resorting to multiple testing, as it is more efficient and empirically we do not observe any performance degradation.

\textbf{Efficiency}. Several modifications can be made to improve the efficiency of Algorithm \ref{a_slime}. At each step when $n$ is increased to $n'$, we can reuse the existing synthetic samples and only generate additional $n' - n$ perturbation points. One may also note that whenever the outer while loop restarts, we conduct repetitive testings for the first several variables entering the model. To achieve better efficiency, each new run can \emph{condition on} previous runs: if a variable enters the LASSO path in the same order as before and has been tested with significant statistics, no additional testing is needed. Hypothesis testing is only invoked when we select more features than previous runs, or in some rare cases, the current iteration disagrees with previous results. \textcolor{black}{In our experiments, we do not implement the conditioning step for implementation simplicity, as we find the efficiency gain is marginal when selecting a moderate size of features.} 


\section{Empirical Studies}\label{empirical}

\textcolor{black}{Rather than performing a broad-scale analysis, we look at several specific cases as illustrations to show the effectiveness of S-LIME in generating stabilized model explanations.} Scikit-learn \citep{scikit-learn} is used for building black box models. Code for replicating our experiments is available at \url{https://github.com/ZhengzeZhou/slime}.


\subsection{Breast Cancer Data}\label{breast}

We use the widely adopted Breast Cancer Wisconsin (Diagnostic) Data Set  \cite{mangasarian1995breast}, which contains 569 samples and 30 features\footnote{\url{https://scikit-learn.org/stable/modules/generated/sklearn.datasets.load_breast_cancer.html}}. A random forests with 500 trees is trained on $80\%$ of the data as the black box model to predict whether an instance is benign or malignant. It achieves around $95\%$ accuracy on the remaining $20\%$ test data. Since our focus is on producing stabilized explanations for a specific instance, we do not spend additional efforts in hyperparameter tuning to further improve model performance. 

Figure \ref{fig:breast_feature} in Section \ref{lime_insta} has already demonstrated the inconsistency of the selected feature returned by original LIME. In Figure \ref{fig:iterations} below, we show a graphical illustration of four LIME replications on a randomly selected test instance, where the left column of each sub figure shows selected features along with learned linear parameters, and the right column is the corresponding feature value for the sample. These repetitions of LIME applied on the same instance have different orderings for the top two features, and also disagree on the fourth and fifth features. 

\begin{figure}[h]
     \centering
     \begin{subfigure}[b]{0.48\linewidth}
         \centering
         \includegraphics[width=\linewidth]{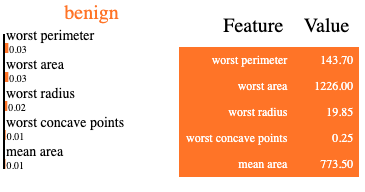}
         \caption{Iteration 1 of LIME}
         \label{fig:i1}
     \end{subfigure}
     \hfill
     \begin{subfigure}[b]{0.48\linewidth}
         \centering
         \includegraphics[width=\linewidth]{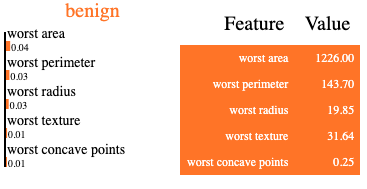}
         \caption{Iteration 2 of LIME}
         \label{fig:i2}
     \end{subfigure}
     \begin{subfigure}[b]{0.48\linewidth}
         \centering
         \includegraphics[width=\linewidth]{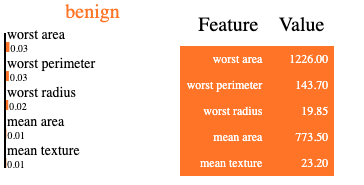}
         \caption{Iteration 3 of LIME}
         \label{fig:i3}
     \end{subfigure}
     \hfill
     \begin{subfigure}[b]{0.48\linewidth}
         \centering
         \includegraphics[width=\linewidth]{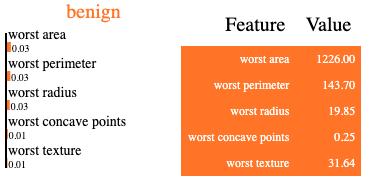}
         \caption{Iteration 4 of LIME}
         \label{fig:i4}
     \end{subfigure}
        \caption{Four iterations of LIME on Breast Cancer Data. The black box model is a random forests classifier with 500 trees. LIME explanations are generated with 1000 synthetic perturbations.}
        \label{fig:iterations}
\end{figure}

To quantify the stability of the generated explanations, we measure the Jaccard index, which is a statistic used for gauging the similarity and diversity of sample sets. Given two sets $A$ and $B$ (in our case, the sets are selected features from LIME), the Jaccard coefficient is defined as the size of the intersection divided by the size of the union:
$$
J(A, B) = \frac{|A \cap B|}{|A \cup B|}.
$$

One disadvantage of the Jaccard index is that it ignores ordering within each feature set. For example, if top two features returned from two iterations of LIME are $A = \{worst\; perimeter,\; worst\; area\}$ and $B = \{worst\; area,\; worst\;perimeter\}$, we have $J(A, B) = 1$ but it does not imply LIME explanations are stable. To better quantify stability, we look at the Jaccard index for the top $k$ features for $k = 1,\ldots,5$. Table \ref{tbl:jaccard} shows the average Jaccard across all pairs for 20 repetitions of both LIME and S-LIME on the selected test instance. We set $n_{max} = 10000$ for S-LIME.

\begin{table}[h]
\centering
\caption{Average Jaccard index for 20 repetitions for LIME and S-LIME. The black box model is a random forests with 500 trees.}
\begin{tabular}{|c|c|c|}
\hline
Position & LIME & S-LIME \\ \hline
1        & 0.61 & 1.0    \\
2        & 1.0  & 1.0    \\
3        & 1.0  & 1.0    \\
4        & 0.66 & 1.0    \\
5        & 0.59 & 0.85   \\ \hline
\end{tabular}
\label{tbl:jaccard}
\end{table}

\begin{figure}[h]
     \centering
     \begin{subfigure}[b]{0.48\linewidth}
         \centering
         \includegraphics[width=\linewidth]{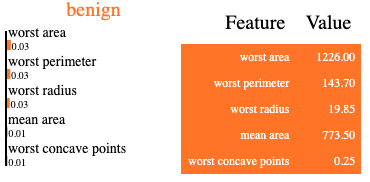}
         \caption{Iteration 1 of S-LIME}
         \label{fig:i1-s}
     \end{subfigure}
     \hfill
     \begin{subfigure}[b]{0.46\linewidth}
         \centering
         \includegraphics[width=\linewidth]{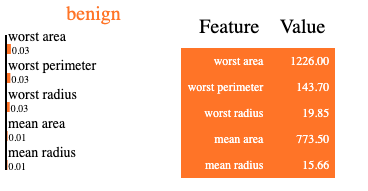}
         \caption{Iteration 2 of S-LIME}
         \label{fig:i2-s}
     \end{subfigure}
        \caption{Two iterations of S-LIME on Breast Cancer Data. The black box model is a random forests classifier with 500 trees.}
        \label{fig:iterations-slime}
\end{figure}

As we can see, for top four positions the average Jaccard index of S-LIME is 1, meaning the algorithm is stable across different iterations. There is some variability in the fifth feature selected, as two features \emph{mean radius} and \emph{worst concave points} have pretty close impact locally. Further increasing $n_{max}$ will make the selection of fifth variable more consistent. Figure \ref{fig:iterations-slime} shows the only two explanations we observed in simulations for S-LIME, where the difference is at the fifth variable.

As a contrast, we've already seen instability for LIME even for the first variable selected. Although LIME consistently selects the same top two and the third feature, there is much variably for the fourth and fifth feature. This experiment demonstrates the stability of S-LIME compared to LIME. In Appendix \ref{app:additional1}, we apply S-LIME on other types of black box models. Stability results on a large cohort of test samples are included in Appendix \ref{app:additional2}.

\subsection{MARS Test Function}

Here we use a modification of the function given in \citep{friedman1991multivariate} (to test the MARS algorithm) as the black box model so we know the underlying true local weights of variables. Let $y = f(\bm{x}) = 10\sin(\pi x_1x_2) + 20(x_3 - 0.05)^2 + 5.2x_4 + 5x_5$, where $\mathcal{X} \sim U([0, 1]^5)$. The test point $\bm{x}$ is chosen to be $(0.51, 0.49, 0.5, 0.5, 0.5)$. We can easily calculate the local linear weights of the five variables around $\bm{x}$ and the expected selection order is $(x_3, x_1, x_2, x_4, x_5)$. Note here the specific choice of parameters in $f(x)$ and the location of test point $\bm{x}$ makes it difficult to distinguish between $x_1, x_2$ and $x_4, x_5$. 

Table \ref{tbl:jaccard2} presents the average Jaccard index for the selected feature sets by LIME and S-LIME, where LIME is generated with 1000 synthetic samples and we set $n_0 = 1000$ and $n_{max} = 10000$ for S-LIME. The close local weights between $x_1, x_2$ and $x_4, x_5$ causes some instability in LIME, as can be seen from the drop in the index at position 2 and 4. S-LIME outputs consistent explanations in this case. 

\begin{table}[h]
\centering
\caption{Average Jaccard index for 20 repetitions for LIME and S-LIME on test point $(0.51, 0.49, 0.5, 0.5, 0.5)$. The black box model is MARS.}
\begin{tabular}{|c|c|c|}
\hline
Position & LIME & S-LIME \\ \hline
1        & 1.0  & 1.0    \\
2        & 0.82 & 1.0    \\
3        & 1.0  & 1.0    \\
4        & 0.79 & 1.0    \\
5        & 1.0  & 1.0      \\ \hline
\end{tabular}
\label{tbl:jaccard2}
\end{table}

\subsection{Early Prediction of Sepsis From Electronic Health Records}\label{sepsis}

Sepsis is a major public health concern which is a leading cause of death in the United States \citep{angus2001epidemiology}. Early detection and treatment of a sepsis incidence is a crucial factor for patient outcomes \citep{reyna2019early}. Electronic health records (EHR) store data associated with each individual’s health journey and have seen an increasing use recently in clinical informatics and epidemiology \citep{solares2020deep, vaid2020federated}. There have been several work to predict sepsis based on EHR \citep{henry2015targeted, futoma2017improved, lauritsen2020early}. Interpretability of these models are essential for them to be deployed in clinical settings.  

We collect data from MIMIC-III \citep{johnson2016mimic}, which is a freely accessible critical care database. After pre-processing, there are 15309 patients in the cohort for analysis, out of which 1221 developed sepsis based on Sepsis-3 clinical criteria for sepsis onset \citep{singer2016third}. For each patient, the record consists of a combination of hourly vital sign summaries, laboratory values, and static patient descriptions. We provide the list of all variables involved in Appendix \ref{app:var_list}. ICULOS is a timestamp which denotes the hours since ICU admission for each patient, and thus is not used directly for training the model.

For each patient's records, missing values are filled with the most recent value if available, otherwise a global average. Negative samples are down sampled to achieve a class ratio of 1:1. We randomly select $90\%$ of the data for training and leave the remaining $10\%$ for testing. A simple recurrent neural network based on LSTM \citep{hochreiter1997long} module is built with Keras \citep{chollet2015keras} for demonstration. Each sample fed into the network has 25 features with 24 timestamps, then goes through a LSTM with 32 internal units with dropout rate 0.2, and finally a dense layer with softmax activation to output a probability. The network is optimized by Adam \citep{kingma2014adam} with an initial learning rate of 0.0001 and we train it for 500 epochs on a batch size of 50. 

The model achieves around 0.75 AUC score on the test set. Note that we do not fine tune the architecture of the network through cross validation. The purpose of this study is not on achieving a superior performance as it usually requires more advanced modeling techniques for temporal data \citep{futoma2017improved, lauritsen2020early} or exploiting missing value patterns \citep{che2018recurrent}. Instead, we would like to demonstrate the effectiveness of our proposed method in reliably explaining a relatively large scale machine learning model applied to medical data.

To deal with temporal data where each sample in the training set is of shape $(n\_timesteps, \, n\_features)$, LIME reshapes the data such that it becomes a long vector of size $n\_timesteps \, \times \, n\_features$. Essentially it transforms the temporal data to the regular tabular shape while increasing the number of features by a multiple of available timestamps. Table \ref{tbl:jaccard_sepsis} presents the average Jaccard index for the selected feature sets by LIME and S-LIME on two randomly selected test samples, where LIME is generated with 1000 synthetic samples and we set $n_0 = 1000$ and $n_{max} = 100000$ for S-LIME. 

LIME exhibits undesirable instability in this example, potentially due to the complex black box model applied and the large number of features ($24 \times 25 = 600$). S-LIME achieves much better stability compared to LIME, although we can still observe some uncertainty in choosing the fifth feature in the second test sample.

\begin{table}[h]
     \caption{Average Jaccard index for 20 repetitions for LIME and S-LIME on two randomly selected test samples. The black box model is a recurrent neural network.}
    \begin{subtable}[h]{0.23\textwidth}
        \centering
        \begin{tabular}{|c|c|c|}
        \hline
        Position & LIME & S-LIME \\ \hline
        1        & 0.37 & 1.0  \\
        2        & 0.29  & 1.0   \\
        3        & 0.33  & 1.0   \\
        4        & 0.25 & 0.89  \\
        5        & 0.26 & 1.0  \\ \hline
        \end{tabular}
        \caption{test sample 1}
        \label{tbl:jaccard_sepsis1}
    \end{subtable}
    \hfill
    \begin{subtable}[h]{0.23\textwidth}
        \centering
        \begin{tabular}{|c|c|c|}
        \hline
        Position & LIME & S-LIME \\ \hline
        1        & 0.31 & 1.0   \\
        2        & 0.24  & 1.0    \\
        3        & 0.19  & 1.0    \\
        4        & 0.17 & 0.96    \\
        5        & 0.18 & 0.78  \\ \hline
        \end{tabular}
        \caption{test sample 2}
        \label{tbl:jaccard_sepsis2}
     \end{subtable}
     \label{tbl:jaccard_sepsis}
\end{table}

Figure \ref{fig:slime_sepsis} below shows the output of S-LIME on two different test samples. We can see that for sample 1, most recent temperatures play an important role, along with the latest pH and potassium values. While for sample 2, latest pH values are the most important ones. 

\begin{figure}[h]
     \centering
     \begin{subfigure}[b]{\linewidth}
     \centering
     \includegraphics[width=\linewidth]{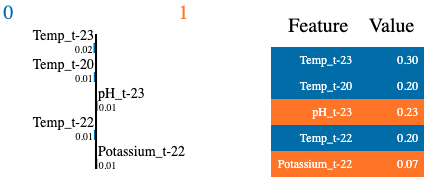}
     \caption{test sample 1}
     \label{fig:slime_sepsis1}
     \end{subfigure}
     \begin{subfigure}[b]{\linewidth}
     \includegraphics[width=\linewidth]{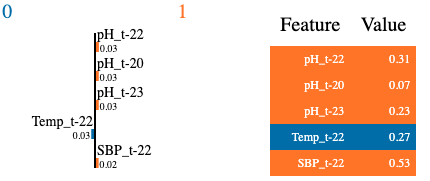}
     \caption{test sample 2}
     \label{fig:slime_sepsis2}
     \end{subfigure}
     \caption{Output of S-LIME for two randomly selected test samples. The black box model is a recurrent neural network.}
    \label{fig:slime_sepsis}
\end{figure}

We want to emphasize that extra caution must be taken by practitioners in applying LIME, especially for some complex problems. The local linear model with a few features might not be suitable to approximate a recurrent neural network built on temporal data. How to apply perturbation based explanation algorithms to temporal data is still an open problem, and we leave it for future work. That being said, the experiment in this section demonstrates the effectiveness of S-LIME in producing stabilized explanations. 

\section{Discussions}\label{conclusion}

An important property for model explanation methods is stability: repeated runs of the algorithm on the same object should output consistent results. In this paper, we show that post hoc explanations based on perturbations, such as LIME, are not stable due to the randomness introduced in generating synthetic samples. Our proposed algorithm S-LIME is based on a hypothesis testing framework and can automatically and adaptively determine the appropriate number of perturbations required to guarantee stability. 


The idea behind S-LIME is similar to \cite{zhou2018approximation} which tackles the problem of building stable approximation trees in model distillation. In the area of online learning, 
\cite{domingos2000mining} uses Hoeffding bounds \citep{hoeffding1994probability} to guarantee correct choice of splits in a decision tree by comparing two best attributes. We should  mention that S-LIME is not restricted to LASSO as its feature selection mechanism. In fact, to produce a ranking of explanatory variables, one can use any sequential procedures which build a model by sequentially adding or removing variables based upon some criterion, such as forward-stepwise or backward-stepwise selection \citep{friedman2001elements}. \textcolor{black}{All of these methods can be stabilized by a similar hypothesis testing framework like S-LIME. }

There are several works closely related to ours. \cite{zhang2019should} identifies three sources of uncertainty in LIME: sampling variance, sensitivity to choice of parameters and variability in the black box model. We aim to control the first source of variability as the other two depend on specific design choices of the practitioners. \cite{visani2020optilime} highlight a trade-off between explanation’s stability and adherence. Their approach is to select a suitable kernel width for the proximity measure, but it does not improve stability \emph{given} any kernel width. In \cite{zafar2019dlime}, the authors design a deterministic version of LIME by only looking at existing training data through hierarchical clustering without resorting to synthetic samples. However, the number of samples in a dataset will affect the quality of clusters and a lack of nearby points poses additional challenges; this strategy also relies of having access to the training data. Most recently, \cite{slack2020much} develop a set of tools for analyzing explanation uncertainty in a Bayesian framework for LIME. Our method can be viewed as a frequentist counterpart without the need to choose priors and evaluate a posterior distribution. 

Another line of work concerns adversarial attacks to LIME. \cite{slack2020fooling} propose a scaffolding technique to hide the biases of any given classifier by building adversarial classifiers to detect perturbed instances. Later, \cite{saito2020improving} utilize a generative adversarial network to sample more realistic synthetic data for making LIME more robust to adversarial attacks. The technique we developed in this work is \emph{orthogonal} to these directions, as  . We also plan to explore other data generating procedures which can help with stability.

\begin{acks}
Giles Hooker is supported by NSF DMS-1712554. Fei Wang is supported by NSF 1750326, 2027970, ONR N00014-18-1-2585, Amazon Web Service (AWS) Machine Learning for Research Award and Google Faculty Research Award.
\end{acks}

\bibliographystyle{ACM-Reference-Format}
\bibliography{reference}

\appendix

\section{Instability with LASSO}\label{app:LASSO}

Instability with LASSO has been studied previously by several researchers. \cite{meinshausen2010stability} introduce stability selection based on subsampling which provides finite sample control for some error rates of false discoveries. \cite{su2018first} find that sequential regression procedures select the first spurious variable unexpectedly early, even in settings of low correlations between variables and strong true effect sizes. \cite{su2017false} further develop a sharp asymptotic trade-off between false and true positive rates along the LASSO path. 

We demonstrate this phenomenon using a simple linear case. Suppose $t = \rho_1x_1 + \rho_2x_2 + \rho_3x_3$, where $x_1$, $x_2$ and $x_3$ are independent and generated from a standard normal distribution $\mathcal{N}(0, 1)$. \textcolor{black}{Note that we do not impose any additional noise in generating the response $y$.} We choose $\rho_1 = 1$, $\rho_2 = 0.75$ and $\rho_3 = 0.7$, such that when one uses LARS to solve LASSO, $x_1$ always enter the model first, while $x_2$ and $x_3$ have closer coefficients and will be more challenging to distinguish. 

We focus on the ordering of the three covariates entering the model. The ``correct" ordering should be $(x_1, x_2, x_3)$. For multiple runs of LASSO with $n = 1000$, we observe roughly $20\%$ of the results have order $(x_1, x_3, x_2)$ instead. Figure \ref{fig:LASSO} below shows two representative LASSO paths. 

\begin{figure}[h]
    \centering
    \begin{subfigure}[b]{0.49\linewidth}
        \centering
        \includegraphics[width=\linewidth]{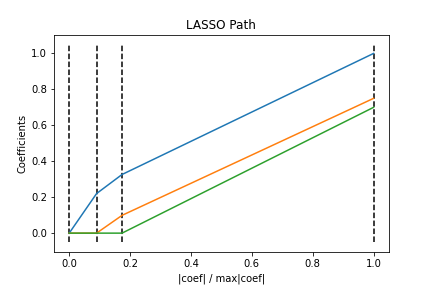}
        \caption{Variable ordering in LASSO path: $(x_1, x_2, x_3)$.}
        \label{fig:LASSO1}
    \end{subfigure}
    \hfill
    \begin{subfigure}[b]{0.49\linewidth}
        \centering
        \includegraphics[width=\linewidth]{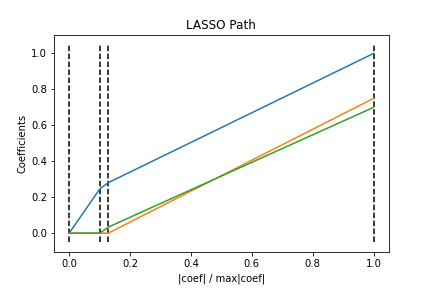}
        \caption{Variable ordering in LASSO path: $(x_1, x_3, x_2)$.}
        \label{fig:LASSO2}
    \end{subfigure}
    \hfill
    
    \caption{Two cases of variable ordering in LASSO path.}
    \label{fig:LASSO}
    
\end{figure}

This toy experiment demonstrates the instability of LASSO itself. Even in this ideal noise-free setting where we have an independent design with Gaussian distribution for the variables, $20\%$ of the time LASSO exhibits different paths due to random sampling. Intuitively, the solutions at the beginning of the LASSO path is overwhelmingly biased and the residual vector contains many of the true effects. Thus some less relevant or irrelevant variable could exhibit high correlations with the residual and gets selected early. $n = 1000$ seems to be a reasonable large number of samples to achieve consistency results, but when applying the idea of S-LIME, the hypothesis testing is always inconclusive at the second step when it needs to choose between $x_2$ and $x_3$. Increasing $n$ in this case can indeed yield significant testing results and stabilize the LASSO paths.

\section{Additional Experiments}\label{app:additional}

\subsection{S-LIME on other model types}\label{app:additional1}

\textcolor{black}{Besides the randomness introduced in generating synthetic perturbations, the output of model explanation algorithms is also dependent on several other factors, including the black box model itself. There may not be a universal truth to the explanations of a given instance, as it depends on how the underlying model captures the relationship between covariates and responses. Distinct model types, or even the same model structure trained with random initialization, can utilize different correlations between features and responses \citep{allen2020towards}, and thus result in different model explanations.}

\textcolor{black}{We apply S-LIME on other model types to illustrate two points:
\begin{itemize}
    \item Compared to LIME, S-LIME can generate stabilized explanations, though for some model types more synthetic perturbations are required.
    \item Different model types can have different explanations for the same instance. This does not imply that S-LIME is unstable or not reproducible, but practitioners need to be aware of this dependency on the underlying black box model when apply any model explanation methods.
\end{itemize}} 

We use support-vector machines (SVM) and neural networks (NN) as the underlying black box models and apply LIME and S-LIME. Basic setups is similar to Section \ref{breast}. For SVM training, we use default parameters\footnote{\url{https://scikit-learn.org/stable/modules/svm.html\#svm-classification}} where rbf kernel is applied. The NN is constructed with two hidden layers, each with 12 and 8 hidden units. ReLU activations are used between hidden layers while the last layer use sigmoid functions to output a probability. The network is implemented in Keras \citep{chollet2015keras}. Both models achieve over $90\%$ accuracy on the test set.

\begin{table}[h]
\centering
\caption{Average Jaccard index for 20 repetitions for LIME and S-LIME. The black box models are SVM and NN.}
\begin{tabular}{|c|c|c|c|c|}
\hline
\multirow{2}{*}{Position} & \multicolumn{2}{c|}{SVM} & \multicolumn{2}{c|}{NN} \\ \cline{2-5} 
                          & LIME       & S-LIME      & LIME      & S-LIME      \\ \hline
1                         & 1          & 1.0         & 0.73      & 1.0         \\ \hline
2                         & 0.35       & 0.87        & 0.87      & 1.0         \\ \hline
3                         & 0.23       & 0.83        & 0.71      & 0.74        \\ \hline
4                         & 0.19       & 1.0         & 0.66      & 1.0         \\ \hline
5                         & 0.18       & 0.67        & 0.55      & 1.0         \\ \hline
\end{tabular}
\label{tbl:additional}
\end{table}

Table \ref{tbl:additional} lists the average Jaccard index across 20 repetitions for each setting on a randomly selected test instance. LIME is generated with 1000 synthetic samples, while for S-LIME we set $n_{max} = 100000$ for SVM and $n_{max} = 10000$ for NN. Compared with LIME, S-LIME achieves better stability at each position. 

\begin{figure}[h]
    \centering
    \begin{subfigure}[b]{\linewidth}
        \centering
        \includegraphics[width=\linewidth]{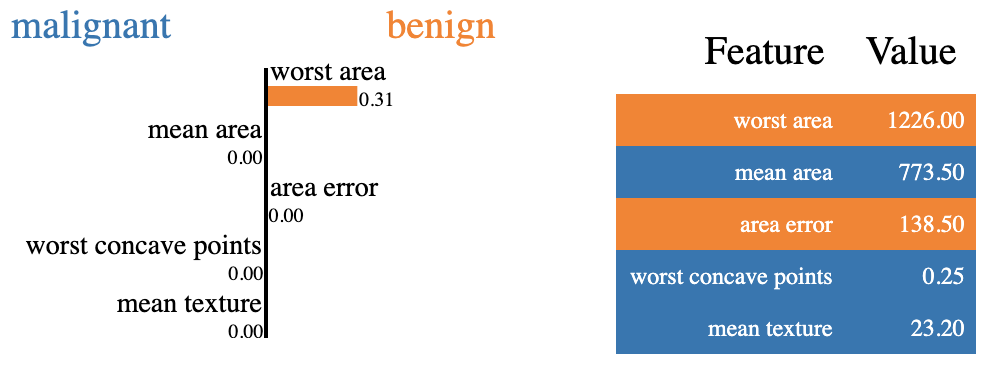}
        \caption{S-LIME on SVM.}
        \label{fig:svm}
    \end{subfigure}
    \hfill
    \begin{subfigure}[b]{\linewidth}
        \centering
        \includegraphics[width=\linewidth]{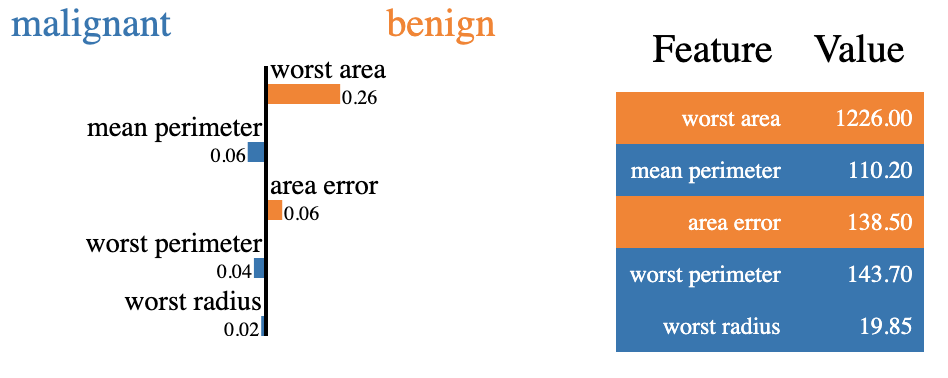}
        \caption{S-LIME on NN.}
        \label{fig:nn}
    \end{subfigure}
    \hfill
    
    \caption{S-LIME on Breast Cancer Data with SVM and NN as black box models.}
    \label{fig:more-slime}
    
\end{figure}

\textcolor{black}{Figure \ref{fig:more-slime} shows the graphical exhibition of the explanations generated by S-LIME for both SVM and NN being the black box models. We can see that they differ in the features selected.}

One important observation is that the underlying black box model also affects the stability of local explanations. For example, the original LIME is extremely unstable for SVM. S-LIME needs a larger $n_{max}$ to produce consistent results. 

\subsection{A large cohort of test samples}\label{app:additional2}

Most of the experiments in this paper are targeted at a randomly selected test sample, which allows us to examine specific features easily. That being said, one can expect the instability of LIME and the improvement of S-LIME to be universal. In this part we conduct experiments on a large cohort of test samples for both Breast Cancer (Section \ref{breast}) and Sepsis (Section \ref{sepsis}) data. 

In each application, we randomly select 50 test samples. For each test instance, LIME and S-LIME are applied for 20 repetitions and we calculate average Jaccard index across all pairs out of 20 as before. Finally, we report the overall average Jaccard index for 50 test samples. The results are shown in Table \ref{tbl:jaccard_50}. LIME explanations are generated with 1000 synthetic samples.  

For Breast Cancer Data, we pick $n_{max} = 10000$ as in Section \ref{breast}. We can see that in general there is some instability from the features selected by LIME, while S-LIME can improve stability. By further increasing $n_{max}$ we may get better stability metrics, but at the cost of computational costs. 

For the sepsis prediction task, LIME performs much worse exhibiting undesirable instability across 50 test samples at all 5 positions. S-LIME with $n_{max} = 100000$ achieves obviously stability improvement. The reason for invoking a larger value of $n_{max}$ is due to the fact that there are 600 features to select from. It is an interesting future direction to see how one can use LIME to explain temporal models more efficiently.

\begin{table}[h]
     \caption{Overall average Jaccard index for 20 repetitions for LIME and S-LIME across 50 randomly chosen test samples.}
    \begin{subtable}[h]{0.23\textwidth}
        \centering
        \begin{tabular}{|c|c|c|}
        \hline
        Position & LIME & S-LIME \\ \hline
        1        & 0.90 & 0.98  \\
        2        & 0.85 & 0.96  \\
        3        & 0.82 & 0.92  \\
        4        & 0.81 & 0.96  \\
        5        & 0.80 & 0.84  \\ \hline
        \end{tabular}
        \caption{Breast Cancer Data}
        \label{tbl:jaccard_bcd_50}
    \end{subtable}
    \hfill
    \begin{subtable}[h]{0.23\textwidth}
        \centering
        \begin{tabular}{|c|c|c|}
        \hline
        Position & LIME & S-LIME \\ \hline
        1        & 0.54 & 1.0   \\
        2        & 0.43 & 1.0 \\
        3        & 0.37 & 0.78 \\
        4        & 0.35 & 0.90   \\
        5        & 0.34 & 0.99  \\ \hline
        \end{tabular}
        \caption{Sepsis Data}
        \label{tbl:jaccard_sepsis_50}
     \end{subtable}
     \label{tbl:jaccard_50}
\end{table}

\newpage

\section{Variables list for Sepsis detection}\label{app:var_list}

\begin{table}[h]
\centering
\caption{Variables list and description for data used in sepsis prediction.}
\begin{tabular}{|l|l|l|}
\hline
\# & Variables  & Description                                              \\ \hline
1  & Age        & age(years)                                               \\ \hline
2  & Gender     & male (1)   or female (0)                                 \\ \hline
3  & ICULOS     & ICU length of stay (hours since ICU admission)           \\ \hline
4  & HR         & hea1t   rate                                             \\ \hline
5  & Potassium  & potassium                                                \\ \hline
6  & Temp       & temperature                                              \\ \hline
7  & pH         & pH                                                       \\ \hline
8  & PaCO2      & partial   pressure of carbon dioxide from arterial blood \\ \hline
9  & SBP        & systolic   blood pressure                                \\ \hline
10 & FiO2       & fraction   of inspired oxygen                            \\ \hline
11 & SaO2       & oxygen   saturation from arterial blood                  \\ \hline
12 & AST        & aspartate   transaminase                                 \\ \hline
13 & BUN        & blood   urea nitrogen                                    \\ \hline
14 & MAP        & mean   arterial pressure                                 \\ \hline
15 & Calcium    & calcium                                                  \\ \hline
16 & Chloride   & chloride                                                 \\ \hline
17 & Creatinine & creatinine                                               \\ \hline
18 & Bilirubin  & bilirubin                                                \\ \hline
19 & Glucose    & glucose                                                  \\ \hline
20 & Lactate    & lactic   acid                                            \\ \hline
21 & DBP        & diastolic   blood pressure                               \\ \hline
22 & Troponin   & troponin   I                                             \\ \hline
23 & Resp       & respiration   rate                                       \\ \hline
24 & PTT        & partial   thromboplastin time                            \\ \hline
25 & WBC        & white   blood cells count                                \\ \hline
26 & Platelets  & platelet   count                                         \\ \hline
\end{tabular}
\end{table}

\end{document}